\documentclass{article}

\usepackage[final]{neurips_2023}
\usepackage{color}
\definecolor{light}{rgb}{0.3, 0.3, 0.3}
\def\light#1{{\color{light}#1}}

\newcommand{\CLS}[1]{$\mathrm{[CLS]}$ #1}

\usepackage[utf8]{inputenc} 
\usepackage[T1]{fontenc}    
\usepackage{hyperref}       
\usepackage{url}            
\usepackage{booktabs}       
\usepackage{amsfonts}       
\usepackage{nicefrac}       
\usepackage{microtype}      
\usepackage{xcolor}         
\usepackage[math]{alp}
\usepackage{graphicx}
\usepackage{wrapfig}
\usepackage{multirow}
\title{On Separate Normalization in Self-supervised Transformers}

\author{%
  Xiaohui Chen \\
  Department of Computer Science\\
  Tufts University\\
  Medford, MA 02155 \\
  \texttt{xiaohui.chen@tufts.edu} \\
  \And
  Yinkai Wang \\
  Department of Computer Science\\
  Tufts University\\
  Medford, MA 02155 \\
  \texttt{yinkai.wang@tufts.edu} \\
  \AND
  Yuanqi Du \\
  Department of Computer Science\\
  Cornell University\\
  Ithaca, NY 14850 \\
  \texttt{yd392@cornell.edu} \\
  \And
  Soha Hassoun \\
  Department of Computer Science\\
  Tufts University\\
  Medford, MA 02155 \\
  \texttt{soha.hassoun@tufts.edu} \\
  \And
  Li-Ping Liu \\
  Department of Computer Science\\
  Tufts University\\
  Medford, MA 02155 \\
  \texttt{liping.liu@tufts.edu} \\
}

\begin{document}

\maketitle

\begin{abstract}
     Self-supervised training methods for transformers have demonstrated remarkable performance across various domains. Previous transformer-based models, such as masked autoencoders (MAE), typically utilize a single normalization layer for both the class token $\mathrm{[CLS]}$ and the tokens. We propose in this paper a new yet simple normalization method that separately normalizes embedding vectors respectively corresponding to normal tokens and the $\mathrm{[CLS]}$ token, in order to better capture their distinct characteristics and enhance downstream task performance. Our empirical study shows that the $\mathrm{[CLS]}$ embeddings learned with our separate normalization layer better encode the global contextual information and are distributed more uniformly in its anisotropic space. When the conventional normalization layer is replaced with a separate normalization layer, we observe an average  2.7\% performance improvement in learning tasks from the image, natural language, and graph domains.
    
\end{abstract}
\section{Introduction}

Transformer models~\citep{vaswani2017attention} have revolutionized natural language processing (NLP)~\citep{devlin2018bert, liu2019roberta} and demonstrated remarkable performances across a wide range of NLP tasks. The significance of transformer models lies in their ability to model context and capture complex linguistic patterns without being constrained by the sequential nature of data. Beyond NLP transformers have further found their successes in areas such as computer vision (CV)~\citep{han2022survey}, speech recognition~\citep{karmakar2021thank}, and recommendation systems~\citep{sun2019bert4rec, gu2020deep, wu2020sse}. Their flexible architecture and ability to capture dependencies have made them adaptable to diverse data modalities in these domains. 

Transformer architectures have been studied extensively from various perspectives such as attention mechanisms, positional encoding~\citep{devlin2018bert}, and normalization techniques. Specifically, layer normalization~\citep{ba2016layer} and batch normalization~\citep{ioffe2015batch} are employed to enhance stability and speed up convergence during training. The literature on transformers also explores parameter initialization~\citep{xu2019lipschitz}, optimization algorithms~\citep{huang2020improving}, regularization techniques~\citep{steiner2021train, zhou2020scheduled}, and improved architectures~\citep{han2021transformer}. This collective research has advanced transformer architectures and their applications in NLP, CV, and other learning domains.

The study of normalization in transformer architectures is motivated by several factors~\citep{xiong2020layer, shen2020powernorm, nguyen2019transformers}. For example, \citet{xiong2020layer} emphasize the importance of the warm-up of the learning rate and the position of layer normalization layers for the purpose of stable training and faster convergence. \citet{shen2020powernorm} investigates the disadvantage of using batch normalization in transformers and proposes power normalization. While most previous works focus on how the normalization layer can be modified to stabilize the training process, it is less understood how the normalization affects the encoding abilities of these embeddings.
\begin{figure}
    \centering
    \includegraphics[width=\textwidth]{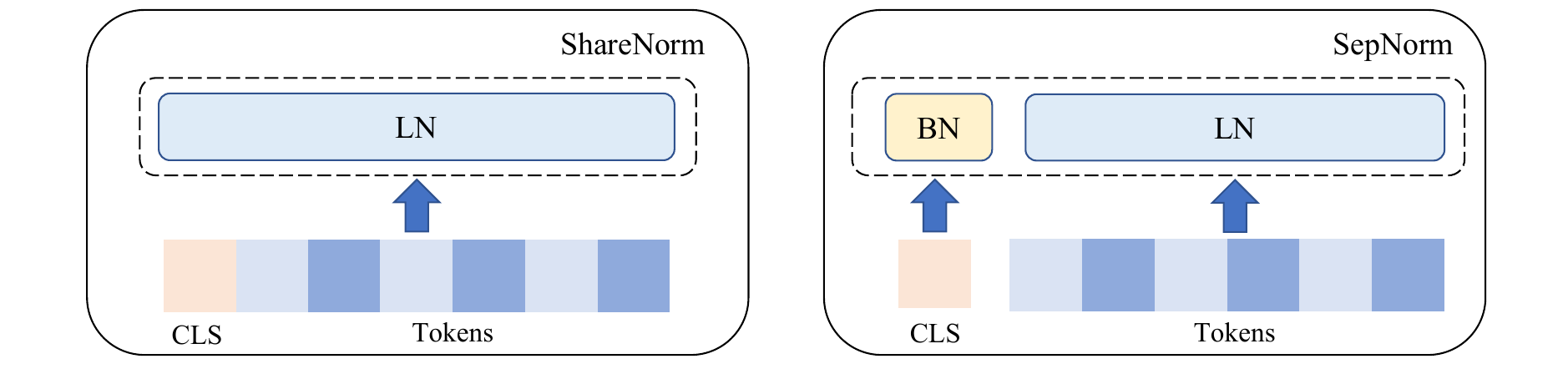}
    \caption{Comparison of the shared normalization (ShareNorm, left) and the proposed separate normalization (SepNorm, right) configurations for token normalization. In the ShareNorm setup, both the \CLS symbol and other tokens are normalized using a single-layer normalization. 
    In the SepNorm setup, normalization is done separately: the \CLS symbol is normalized through batch normalization, while other tokens are normalized via layer normalization.
    }
    \vspace{-1em}
    \label{fig:dualnorm}
\end{figure}

In self-supervised transformers, the \CLS symbol is frequently used as a global representation for various downstream tasks~\citep{devlin2018bert,he2022masked}. Often, the normalization applied to the \CLS symbol is shared with the rest of the tokens in the sequence, which we term it as Shared Normalization (ShareNorm). Given that the \CLS symbol plays a special role in representation learning, a natural question is whether we should treat it separately in the normalization operation. Driven by this question, our research first scrutinizes the behavior of the current shared normalization in transformers, particularly the properties of the \CLS embedding and its influence on downstream task performance. Subsequently, we propose a replacement of ShareNorm with Separate Normalization (SepNorm), the latter of which employs distinct normalization operations for the \CLS symbol and the token features, as depicted in Figure \ref{fig:dualnorm}. Through extensive analysis, we demonstrate that  \CLS embeddings learned using ShareNorm have the issue of dimensional collapse, which cannot be rectified even by enforcing uniformity~\citep{wang2020understanding}. However, the straightforward substitution of SepNorm for ShareNorm substantially mitigates this issue. We empirically validate the effectiveness of SepNorm in tasks from the image, text, and graph domains, demonstrating the universal advantage of the proposed SepNorm.

\section{Background}
\subsection{Pretraining Transformers with the \CLS symbol}

Unsupervised pretraining of a transformer-based model~\citep{vaswani2017attention} is widely investigated in many domains, including NLP, computer vision (CV), and graphs.

\paragraph{Pretraining BERT for NLP.} In NLP, \citet{devlin2018bert} first develop the BERT model by pretraining a transformer-based network by performing the following two tasks -- masked language modeling and next sentence prediction. During pretraining, BERT takes a pair of sentences $(\bx,\by)$, which are represented as a special sequence
\begin{align}
\bs=\big(\mathrm{[CLS]}, \bx,  \mathrm{[SEP]},\by \big).
\end{align}
Here $\mathrm{[SEP]}$ is a special token that separates the two sentences. A fraction (e.g., 15\%) of the tokens in $\bx$ and $\by$ are randomly replaced by a special symbol $\mathrm{[MASK]}$. The first task in BERT is to predict the original tokens replaced by $\mathrm{[MASK]}$ with cross-entropy loss. The second task is to predict whether $\by$ is the next sentence following $\bx$, and the decision is made by classifying the final embedding of the \CLS symbol. After pretraining, the representation in the \CLS is usually used for sentence-level downstream tasks such as sentiment analysis~\citep{medhat2014sentiment}.

\paragraph{Pretraining MAE for CV.}
The Vision Transformer (ViT)~\citep{dosovitskiy2020image} applies the transformers to computer vision tasks. In ViT, an image is usually voxelized into $16 \times 16$ patches, which are then flattened into a sequence of 256 tokens and fed into the ViT. \citet{he2022masked} proposes a self-supervised training scheme, Masked Autoencoder (MAE), for the ViT architecture. A training image has 75\% of its patches masked. The MAE feeds Tokens of unmasked patches as well as a \CLS token into the encoder and gets the representations for these tokens. Then the decoder tries to reconstruct the original image by minimizing the mean square error (MSE). Only the encoder will be used for downstream tasks after pretraining. The \CLS symbol is treated as the class token for linear probing and fine-tuning in the downstream tasks.

\paragraph{Pretraining Graphormer for molecule discovery.} 
Graphormer~\citep{ying2021transformers} is a transformer-based model designed for graph representation learning tasks. It is used to predict the property of a graph rather than a node or edge. Specifically, Graphormer introduces a new symbol $\mathrm{[VNode]}$ as a node connecting to all original graph nodes. Then the vector learned for $\mathrm{[VNode]}$ represents the global information of the entire graph. The mechanism of $\mathrm{[VNode]}$ is similar to the \CLS symbol in BERT and MAE.

In typical applications of transformers, the \CLS symbol is not a natural data token. It summarizes other tokens to capture global information, which is especially useful in downstream tasks. For these reasons, we argue that it should be treated differently in normalization operations. 

\subsection{Normalization Layers in Transformers}

Given that transformers are initially proposed for NLP tasks, layer normalization (LN)~\citet{ba2016layer} is typically the normalization method of choice~\citep{xiong2020layer}. LN normalizes across feature dimensions and is independent of the sequence length and the batch size. For any features $\bh\in\mathbb{R}^d$, the LN has the following computation:
\begin{align}
    \mathrm{LN}(\bh) = \bgamma \odot \frac{\bh-\mu}{\sigma} +\bbeta,\quad\mu = \frac{1}{d}\sum_{i=1}^d h_i,\quad\sigma = \sqrt{\frac{1}{d}\sum^{d}_{i=1} \big(h_i-\mu\big)^2}.
\end{align}
Here $h_i$ is the $i$-th dimension of $\bh$, $\odot$ represents element-wise multiplication, and $\bgamma,\bbeta\in\mathbb{R}^d$ are scale and bias parameters, respectively. In a transformer, all tokens, including special tokens, such as $\mathrm{[CLS]}$ and $\mathrm{[SEP]}$, are all treated equally and share the same LNs. 

Batch Normalization (BN)~\citep{ioffe2015batch} works by normalizing the input data to have zero mean and unit variance along the batch dimension, followed by an affine transformation to scale the result using gamma and beta parameters. BN normalizes a given vector $\bh$ as:
\begin{align}
    \mathrm{BN}(\bh) =  \bgamma \odot \frac{\bh-\bmu_B}{\bsigma_B} +\bbeta.
\end{align}
Here $\bmu_B, \bsigma_B^2 \in \mathbb{R}^d$ are the running statistics (mean and variance) maintained by the BN. The running mean and variance are updated during training after each batch. They are usually calculated as an exponential moving average of the batch mean and variance. BN is widely adopted in CV but leads to significant performance degradation when naively used in NLP.




\subsection{Uniformity of the Learned Representations}

The dimensional collapse in self-supervised representation learning is a common phenomenon where the embedding vectors only span a lower-dimensional subspace~\citep{jing2021understanding} of the entire vector space. This means that the model fails to capture data patterns with full power and instead collapses to a simpler representation. Contrastive methods~\citep{oord2018representation,chen2020simple} have been one of the standard approaches to address this problem. Specifically, \citet{wang2020understanding} propose the \textit{uniformity} metric (loss) to quantify the degree of dimensional collapse. Given a set of representation vectors $\{\bh_1,\ldots,\bh_N\}$ from a dataset of size $N$,  the uniformity metric $\calL_\calU $ is computed as follows:
\begin{align}
    \calL_\calU = \log \frac{1}{N(N-1)/2}\sum_{\substack{n=1,\\m=n+1}}^{N,N} \mathrm{exp}^{-2\big\|\frac{\bh_n}{\|\bh_n\|} - \frac{\bh_m}{\|\bh_m\|}\big\|^2}.\label{eq:unif}
\end{align}
If the distribution of the representation is perfectly uniform, then the numerical value of $\calL_\calU$ will converge to -4 as the dimension of $\bh$ increases to infinity~\citep{wang2020understanding}. 

In self-supervised transformers, the uniformity of the representation is also taken into consideration by some works. For example, \citet{gao2021simcse} finetune the pretrained BERT model using the InfoNCE loss~\citep{oord2018representation}, and \citet{zhang2022mask} jointly train the MAE loss along with uniformity loss.

\section{Approach}
\subsection{Separate Normalization}

We present SepNorm, a normalization scheme that separately normalizes embeddings of the \CLS symbol and embeddings of other tokens. In this work, we focus on the exploration of combinations of BN and LN for the two separate normalization channels. 

For instance, if we apply BN to the \CLS symbol and LN to other tokens, the learnable parameters are structured as $g_1=(\bgamma_1, \bbeta_1)$ and $g_2 =(\bgamma_2, \bbeta_2)$. Let $\bH\in\mathbb{R}^{L\times d}$ represent the feature sequence, where $L$ denotes the sequence length, and $d$ is the feature dimension. Assume embedding $\bH_0$ in the first position corresponds to the \CLS symbol. The normalization process is as follows:
\begin{align}
    \bH' = \big(\mathrm{BN}(\bH_0;g_1),\mathrm{LN}(\bH_{1};g_2),\ldots,\mathrm{LN}(\bH_{L};g_2)\big),
\end{align}
where $\bH'$ denotes the normalized features. We can also run separate normalization with one of the three other combinations:  
\begin{align*}
    \bH' &= \big(\mathrm{BN}(\bH_0;g_1),\mathrm{BN}(\bH_{1};g_2),\ldots,\mathrm{BN}(\bH_{L};g_2)\big), \\
    \bH' &= \big(\mathrm{LN}(\bH_0;g_1),\mathrm{BN}(\bH_{1};g_2),\ldots,\mathrm{BN}(\bH_{L};g_2)\big), \\
    \bH' &= \big(\mathrm{LN}(\bH_0;g_1),\mathrm{LN}(\bH_{1};g_2),\ldots,\mathrm{LN}(\bH_{L};g_2)\big).
\end{align*}

Separate normalization allows the \CLS features to be encoded distinctly from other tokens. 

\begin{figure}[t]
    \centering
    \small
    \begin{tabular}{ccc}
    \multicolumn{3}{c}{\begin{tabular}{ccc}
         \includegraphics[width=0.42\textwidth]{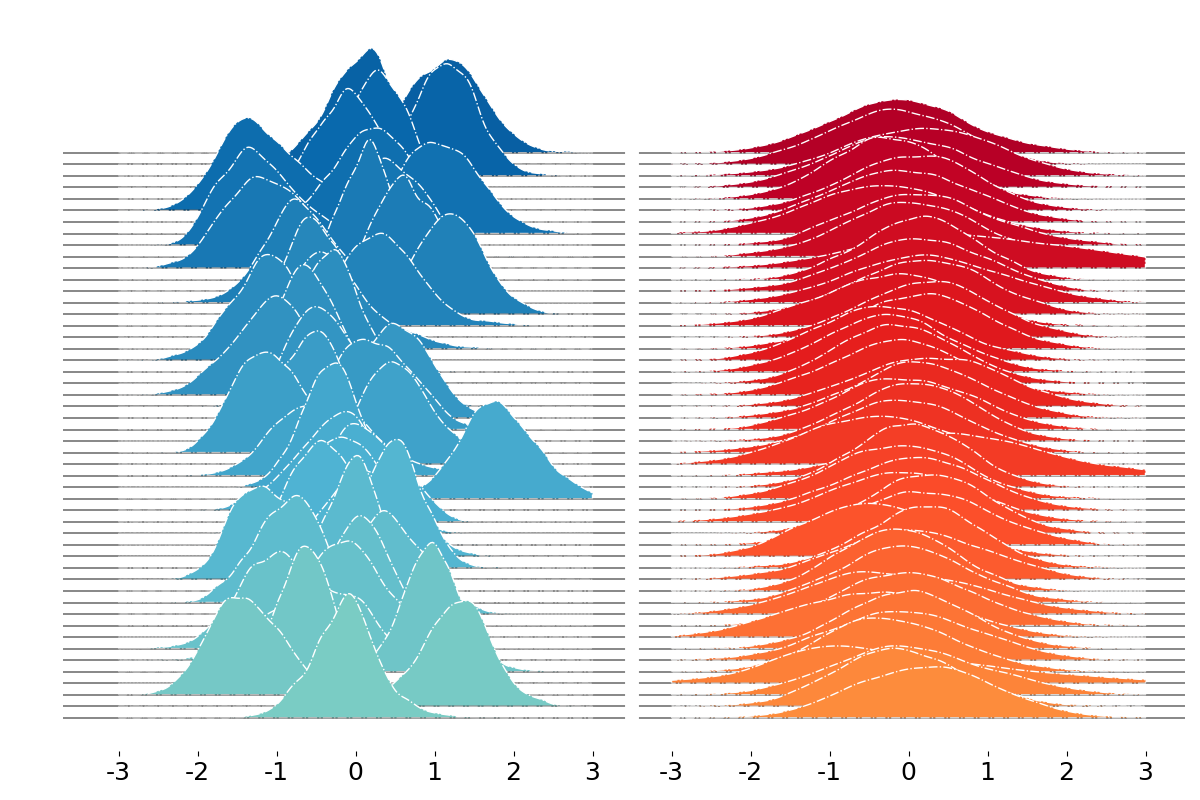}& &\includegraphics[width=0.42\textwidth]{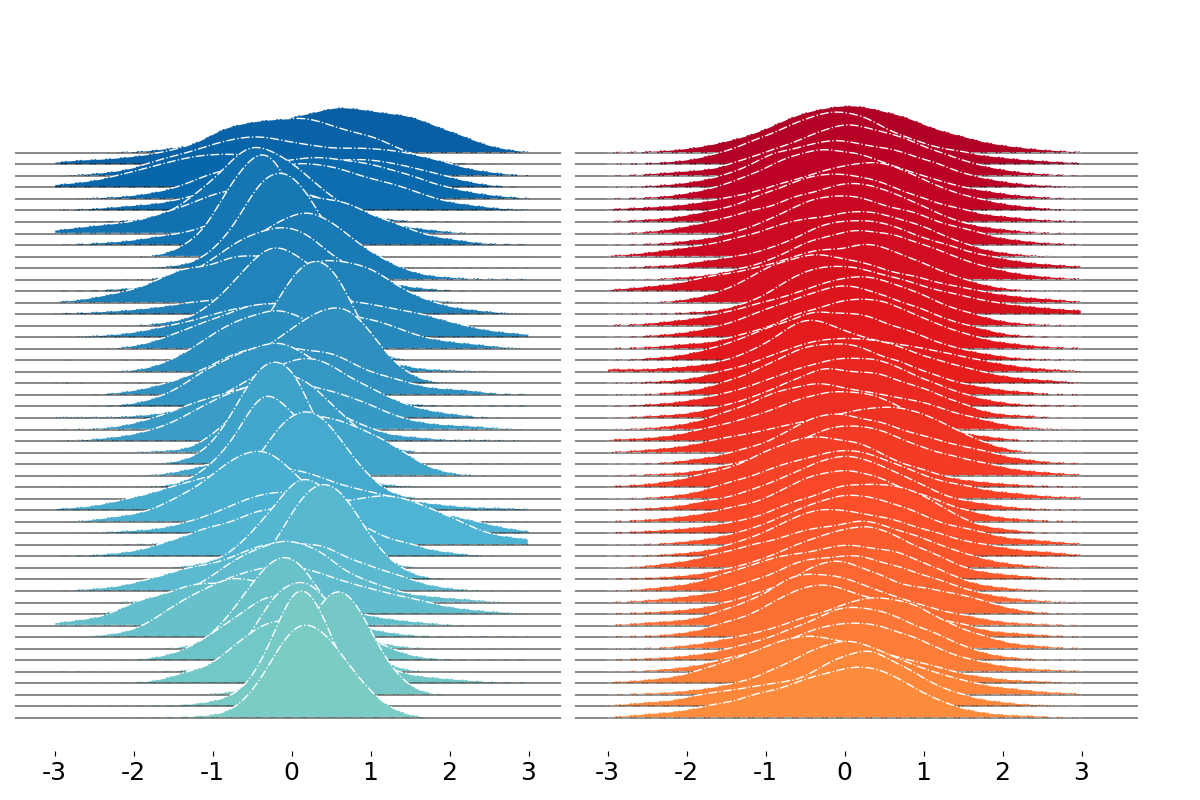}\\
         (a) Shared Normalization (ShareNorm) && (b) Separate Normalization (SepNorm) 
    \end{tabular}}\\\\
    \end{tabular}
    \caption{The effect of SepNorm on feature distributions. Each subplot shows the distributions of the first 50 feature dimensions: \CLS features are in blue, and other tokens' features are in red. The \CLS features of ShareNorm are more concentrated around the mean and the mean deviates more from the zero centers, while in SepNorm, the \CLS distribution is more centered and flattened. 
    }
    \vspace{-1em}
    \label{fig:dist_landscape}
\end{figure}

As a comparison, the \CLS token's embedding and other tokens' embeddings interfere with each other in a shared normalization structure.  With ShareNorm, the update directions of the LN parameters $\{\bgamma,\bbeta\}$ are primarily driven by the embeddings of normal tokens. Below is the gradient calculation for these parameters, 
\begin{align}
    &\frac{\delta\calL}{\delta \gamma_i} = \sum_{l=1}^L \frac{\delta\calL}{\delta \tilde{\bH}_{l, i}}\tilde{\bH}_{l, i}, \quad \frac{\delta\calL}{\delta \beta_i} = \sum_{l=1}^L \frac{\delta\calL}{\delta \tilde{\bH}_{l, i}},
    \\
    &\text{where}~\tilde{\bH}_{l, i} = \frac{\bH_{l,i}-\mu_l}{\sigma_l}, \mu_l = \frac{1}{d}\sum_{i=1}^d\bH_{l,i},\sigma_l = \sqrt{\frac{1}{d}\sum^{d}_{i=1} \big(\bH_{l,i}-\mu_l\big)^2}.
\end{align}
We see the summation in the gradient calculation is dominated by normal tokens given that the number of normal tokens is typically a large number. Given the potentially diverse characteristics (i.e., mean and scale) of feature distributions, it might be challenging for normalization parameters to accommodate both token types simultaneously. Moreover, mapping two types of token features into the same sphere may also mix the signal of \CLS tokens with other tokens.  Figure~\ref{fig:dist_landscape}(a, b) demonstrates this phenomenon in the scenario where both token types utilize a ShareNorm and how using SepNorm mitigates this effect.

\subsection{Encourage the Uniformity of the \CLS Embeddings via a Constrastive Term}

We further relate SepNorm with the uniformity of embeddings. Higher uniformity values indicate that embeddings can better exploit the space to store information. Contrastive methods often employ negative instances to encourage uniformity. In particular, we incorporate SepNorm into transformers trained with U-MAE~\citep{zhang2022mask}, which uses a constrastive term to promote uniformity of features. 

The U-MAE explicitly adds a uniformity loss term $\calL_{\mathrm{unif}}$ to the training objective to encourage uniformity of \CLS embeddings. 
\begin{align}
    \calL_{\mathrm{U\mbox{-}MAE}} = \calL_\mathrm{MAE} + \lambda \calL_{\mathrm{unif}}, ~ \mbox{ with }
    \calL_{\mathrm{unif}} = \E{i}{\E{j}{\bh^\top_{\mathrm{CLS}, i} \, \bh_{\mathrm{CLS}, j}}} 
\end{align}
Here $\calL_\mathrm{MAE}$ is the MAE training objective. The two indices $i$ and $j$ represent two sequences within the same batch. \CLS embeddings $\bh_{\mathrm{CLS}, i}$ and $\bh_{\mathrm{CLS}, j}$, which are respectively for the two sequences, are obtained from our SepNorm during the transformer calculation. By minimizing $\calL_{\mathrm{unif}}$, \CLS features tend to be different from each other.

\section{Experiments}

We examine the effectiveness of the proposed SepNorm component in three domains: CV, NLP, and graphs. We then further investigate how the ShareNorm and SepNorm affect the uniformity of the \CLS embeddings.

\subsection{Computer Vision}
\paragraph{Datasets.} We investigate the model performance on the four image datasets: STL10~\citep{coates2011analysis}, FGVC Aircraft~\citep{maji2013fine}, Street View House Numbers (SVHN)~\citep{netzer2011reading}, and Oxford 102 Flowers~\citep{nilsback2008automated}. All four datasets are for classification tasks. We follow the train/test split provided in the papers introducing the datasets. We report top-1 and top-5 accuracy for all datasets.

\paragraph{Vision transformers (ViT) and MAE.} We choose Vision Transformer (ViT)~\citep{dosovitskiy2020image} as our feature extractor for all datasets. To pretrain the ViT, we adopt the MAE training scheme~\citep{he2022masked}. We follow MAE and use a 75\% masking ratio on input image. During the downstream tasks, we use the embeddings of the \CLS token to predict the class labels.

\paragraph{Experiment setup.}
We follow the setup in \citet{he2022masked} to pretrain and evaluate the ViT. For pertaining, we train the ViT for 4000 epochs. For linear probing, we freeze the encoder's weight and train the last layer on the specific datasets for 2000 epochs. We use a batch size of 512 for pretraining and a batch size of 128 for linear probing.

\begin{figure}[t]
    \centering
    \begin{tabular}{cc}
      \includegraphics[width=0.35\textwidth]{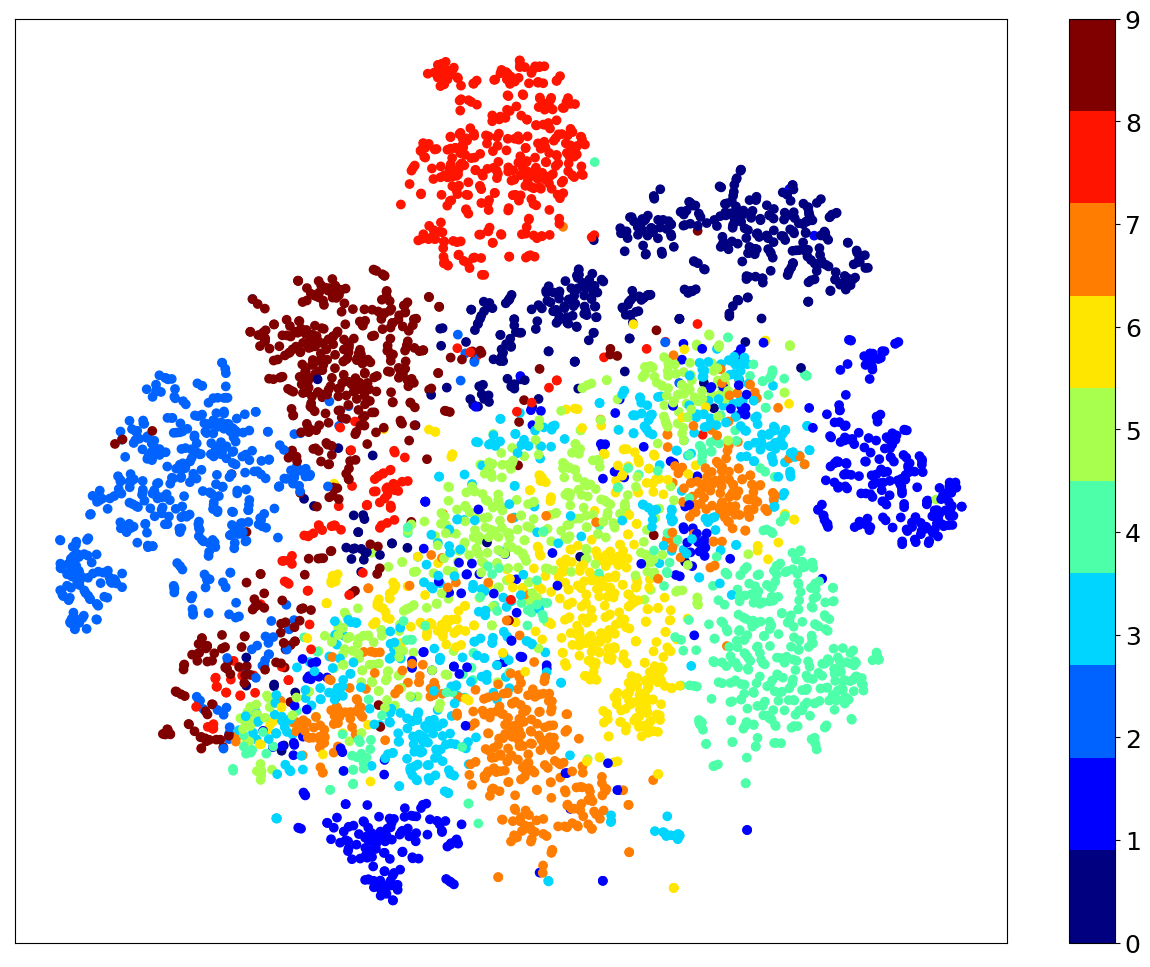}&\includegraphics[width=0.35\textwidth]{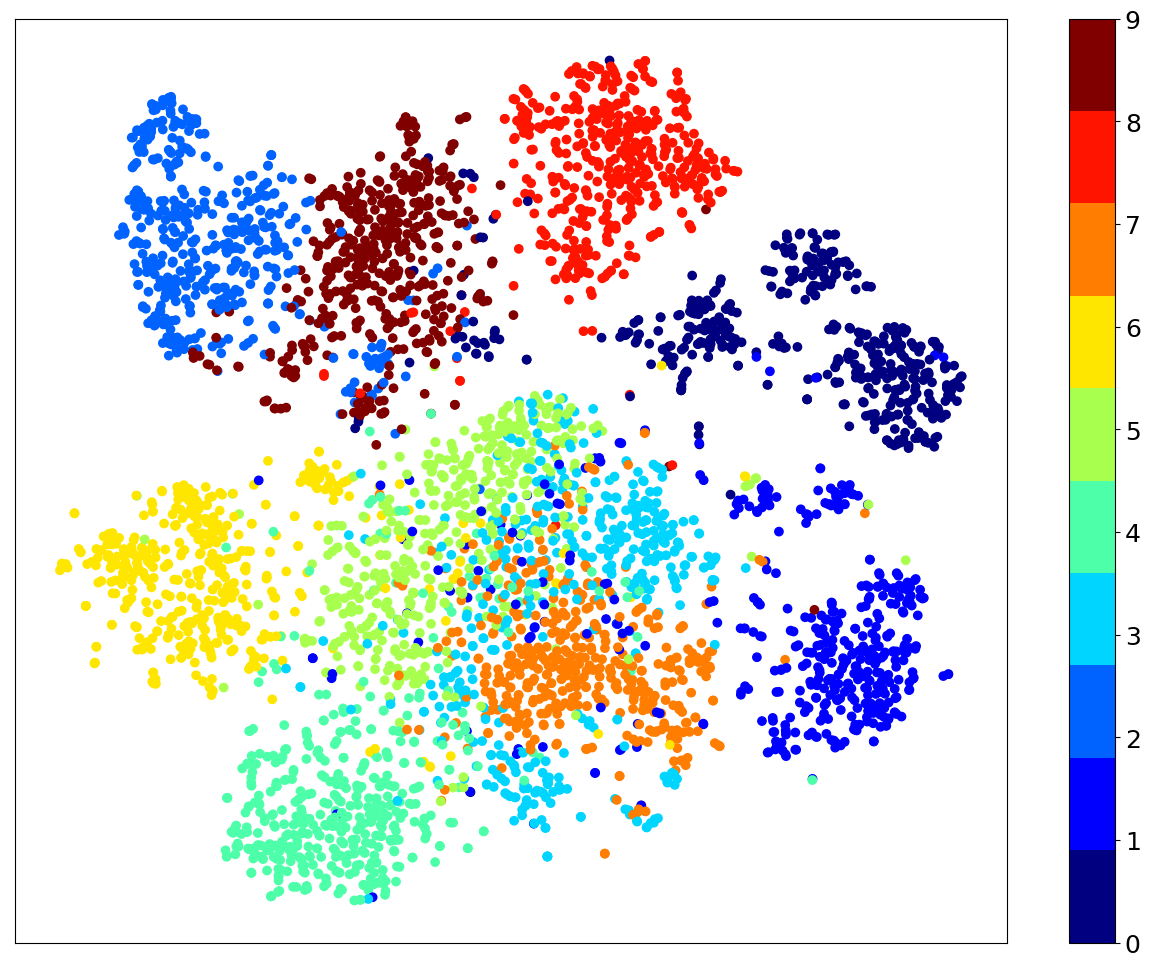}\\
      ShareNorm(LN)& SepNorm(BN+BN)
    \end{tabular}
    \caption{t-SNE visualization of representations learned from the STL-10 dataset.}
    \label{fig:t-sne}
\end{figure}

\paragraph{Experiment results.}
The results presented in Table~\ref{tab:CV_result} demonstrate the performances of our model and the baseline model. Our model consistently outperforms the baseline across multiple datasets, indicating its effectiveness in image classification tasks.
\begin{table}[t]
    \centering
    \small
    \begin{tabular}{@{}lcccccccc@{}}\toprule
         &  \multicolumn{2}{c}{STL10} &\multicolumn{2}{c}{Aircraft} & \multicolumn{2}{c}{SVHN}& \multicolumn{2}{c}{Flower}\\
         &  ACC@1 & ACC@5&ACC@1 & ACC@5&ACC@1 & ACC@5&ACC@1 & ACC@5\\\midrule
        MAE & 92.01 & 99.5 & 52.54 & 84.16 &88.97 & 99.13&27.63& 53.73\\
        + SepNorm & \textbf{93.84} & \textbf{99.7} &\textbf{59.02} & \textbf{86.65} &\textbf{89.18} &\textbf{99.21} & \textbf{32.51} & \textbf{60.92}\\\bottomrule\\
    \end{tabular}
    \caption{Comparison of linear probing performance of ShareNorm and SepNorm across 4 image classification datasets when the $\text{ViT}_\text{\light{base}}$ is pretrained with MAE.}
    \label{tab:CV_result}
\end{table}
In the STL-10 dataset, our approach achieves the top-1 accuracy of 93.84\% and the top-5 accuracy of 99.7\%, higher than the baseline's respective accuracies of 92.01\% and 99.5\%. Similar improvements are observed in the Aircraft, SVHN, and Flower datasets, where our model consistently outperforms the baseline in both top-1 and top-5 accuracies. These results demonstrate the effectiveness of SepNorm in enhancing image classification performance. We also visualize the embeddings of ShareNorm and SepNorm using t-SNE in Figure~\ref{fig:t-sne}. Compared with ShareNorm, SepNorm provides embeddings that have better separation among different classes.

\subsection{Natural Language Processing}

\paragraph{Datasets.}
We evaluated our approach using the STS dataset, which comprises seven semantic textual similarity (STS) tasks. These tasks, including STS 2012-2016 \citep{agirre2012semeval, agirre2013sem, agirre2014semeval, agirre2015semeval, agirre2016semeval}, STS Benchmark \citep{cer2017semeval}, and SICK-Relatedness \citep{marelli2014sick}. We also evaluate our method on multiple transfer tasks, including MR~\citep{pang2005seeing}, CR~\citep{hu2004mining}, SUBJ~\citep{pang2004sentimental}, MPQA~\citep{wiebe2005annotating}, SST-2~\citep{socher2013recursive}, TREC~\cite{voorhees2000building}, and MRPC~\citep{dolan2005automatically}. Following the evaluation settings of SimCSE \citep{gao2021simcse}, we use Spearman's correlation coefficient as the evaluation metric. 

\paragraph{BERT and RoBERTa.}
We conduct our study with pretrained checkpoints of BERT (uncased)~\citep{devlin2018bert} and RoBERTa (cased)~\citep{liu2019roberta}, instead of training them from scratch. Using pretrained models is common in this research field~\citep{gao2021simcse} because the findings are compatible with the common practice of finetuning pretrained models in actual learning tasks. This strategy also saves significant training time and computational resources, allowing us to extend the study to more learning tasks. 

\paragraph{Experiment setup.}
We follow the experiment setup in \citet{gao2021simcse} and further finetune the BERT and RoBERTa models on English Wikipedia. We evaluate the models using established STS tasks and employ standard evaluation metrics such as Spearman's correlation.

\begin{table}[h]
\centering
\small
\begin{tabular}{@{}llcccccccc@{}}
\toprule
&& STS12 & STS13 & STS14 & STS15 & STS16 & STS-B & SICK-R & Avg.\\\midrule
&\multicolumn{9}{c}{Unsupervised Training}\\\midrule
$\text{BERT}_\text{\light{base}}$ & ShareNorm& 65.28 & 78.82 & 69.65 & 79.02 & 77.21 & 76.4 & \textbf{71.74} & 74.04\\
&SepNorm & \textbf{67.01} & \textbf{82.16} & \textbf{72.48} & \textbf{81.38} & \textbf{79.11} & \textbf{77.56} & 71.36&\textbf{75.87}\\
\midrule
$\text{RoBERTa}_\text{\light{base}}$& ShareNorm&\textbf{68.25}&81.24&72.78&{81.38}&\textbf{80.31}&79.83&68.16&76.00\\
&SepNorm & 66.63 &\textbf{82.40}&\textbf{74.47}&\textbf{82.39}&\textbf{80.44}&\textbf{81.14}&\textbf{69.44}&\textbf{76.70}\\\midrule 
&\multicolumn{9}{c}{Supervised Training}\\\midrule
$\text{BERT}_\text{\light{base}}$ & ShareNorm& \textbf{77.72}&
81.07&
\textbf{78.97}&
\textbf{85.15}&
\textbf{82.00}&
82.36&
\textbf{79.74}&
81.00 \\
&SepNorm & 75.32&
\textbf{84.41}&
\textbf{79.94}&
84.91&
80.87&
\textbf{83.63}&
\textbf{79.61}&
\textbf{81.23}
\\
\midrule
$\text{RoBERTa}_\text{\light{base}}$& ShareNorm& 
\textbf{77.38}&
80.87&
78.72&
84.02&
\textbf{82.56}&
83.08&
78.25&
80.70
\\
&SepNorm & 
75.80&
\textbf{84.94}&
\textbf{80.33}&
\textbf{85.51}&
82.11&
\textbf{84.88}&
\textbf{79.72}&
\textbf{81.90}\\\midrule 
 && MR & CR & SUBJ & MPQA & SST2 & TREC & MRPC & Avg.\\
\midrule&\multicolumn{9}{c}{Transfer Learning}\\\midrule
$\text{BERT}_\text{\light{base}}$ & ShareNorm & \textbf{82.78} & 88.79 & \textbf{94.69} & \textbf{89.86} & \textbf{87.94} & \textbf{84.44} & \textbf{75.99} & \textbf{86.36} \\
&SepNorm & \textbf{82.82} & \textbf{89.08} & 94.30 & 89.70 & \textbf{87.97} & {83.88} & 75.21 & 86.14 \\
\midrule
$\text{RoBERTa}_\text{\light{base}}$ & ShareNorm & 84.45 & \textbf{91.50} & 93.94 & \textbf{89.45} & 90.96 & 86.80 & \textbf{76.13} & 87.61 \\
&SepNorm & \textbf{85.11} & \textbf{91.56} & \textbf{94.30} & \textbf{89.43} & \textbf{91.66} & \textbf{90.96} & {75.58} & \textbf{88.37}
\\\bottomrule 
\end{tabular}
\vspace{0.5em}
\caption{Sentence embedding performance on STS tasks and transfer tasks.}
\vspace{-2em}
\label{tab:NLP_result}
\end{table}

\paragraph{Experiment results.}
The experiment results presented in Table~\ref{tab:NLP_result} highlight the performance of our model compared to the SimCSE baseline on NLP tasks. With the SepNorm layer, $\text{BERT}_\text{\light{base}}$ and $\text{RoBERTa}_\text{\light{base}}$ achieve overall higher average accuracy compared to ShareNorm's average accuracy. Only in the transfer learning tasks, SepNorm works slightly worse than ShareNorm in $\text{BERT}_\text{\light{base}}$, but the difference is marginal.

\subsection{Prediction of Molecule Properties}
\paragraph{Datasets.} We conducted experiments using the ZINC dataset~\citep{irwin2005ZINC}, which contains approximately 250,000 molecular graphs. The task is to predict the properties of molecules from their graphs. We use a subset of 12,000 molecular graphs, as recommended by the benchmarking methodology outlined in~\citep{dwivedi2020benchmarking}, so that our results are comparable with other studies. Despite being smaller, the subset retained sufficient diversity and complexity for effective evaluation. We also the MolHIV dataset from the OGB~\citep{hu2020open} collection, which is widely used for training and evaluating graph-based models in molecular property prediction tasks.

\paragraph{Graphormer.}  We use Graphormer~\citep{ying2021transformers} as the transformer backbone to construct the predicting model. 
To obtain graph-level information, Graphormer adds a special node [VNode] to the graph and connects it to all normal graph nodes. The embedding of [VNode] is a summary of the entire graph and will be used in downstream classification tasks. The special node [VNode] serves the same purpose as the \CLS token in traditional Transformer models. Graphormer has used three encodings to enhance the transformer's learning ability: centrality encoding captures node importance, spatial encoding considers spatial relations, and edge encoding incorporates edge features.

\begin{wraptable}[9]{r}{7.2cm}
    \vspace{-5mm}
    \begin{tabular}{@{}lccc@{}}\toprule
        Dataset & ZINC & ZINC (subset) & MolHIV\\\midrule
        Metrics &\multicolumn{2}{c}{Mean absolute error$\downarrow$} & AUC$\uparrow$\\\midrule
        Graphormer & 0.069  & 0.164& 73.36\%\\
        + SepNorm & \textbf{0.052}& \textbf{0.144} &  \textbf{75.64\%}\\\bottomrule
    \end{tabular}
    \caption{A comparison of ShareNorm and SepNorm in three tasks of graph property prediction.}
    \label{tab:Graph_result}
    \vspace{0mm}
\end{wraptable}
\paragraph{Experiment setup.}
We strictly follow Graphormer~\citep{ying2021transformers} in terms of the model architecture, hyperparameters, and training strategies. We replaced the ShareNorm in Graphormer with SepNorm to investigate the effectiveness of the proposed component. 
We evaluate the pretrained model on a broad class of graph-level prediction tasks. We report the mean absolute error for the ZINC and ZINC (subset) datasets and the area under the curve (AUC) for the MolHIV dataset.

\paragraph{Experiment results.}
Table~\ref{tab:Graph_result} shows the performances of our model and the Graphormer baseline. For the ZINC datasets, Graphormer with SepNorm achieves a significantly lower mean absolute error compared to that with ShareNorm. On the MolHIV dataset, SepNorm also improves the AUC to 75.64\%, compared with ShareNorm's AUC of 73.36\%. These results are strong evidence that the embeddings of the [VNode] can better summarize the properties of the entire graph and thus give superior performance on downstream tasks.

\subsection{Uniformity Analysis}

In this section, we investigate how, under both non-contrastive and contrastive training methods, ShareNorm and SepNorm respectively affect the uniformity of learned embeddings and further classification performances.

\begin{figure}[t]
    \centering
    \small
    \begin{tabular}{ccc}
    \includegraphics[width=0.305\textwidth]{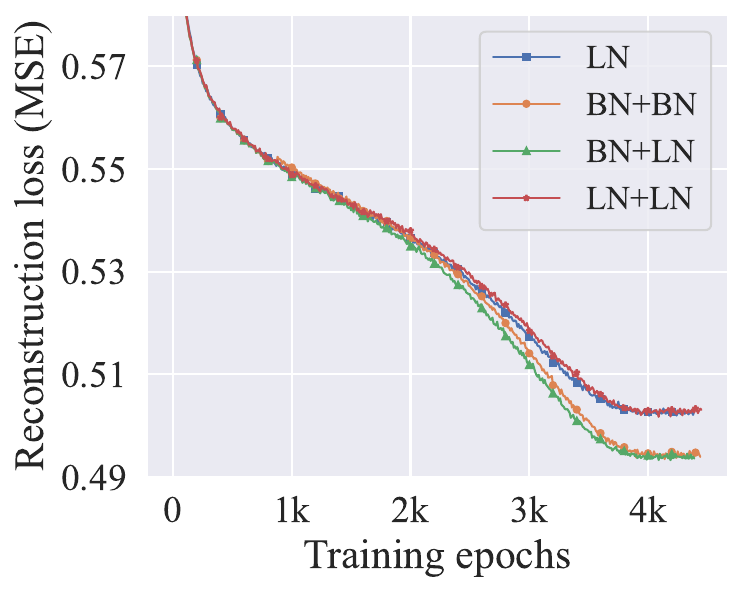} &\includegraphics[width=0.3\textwidth]{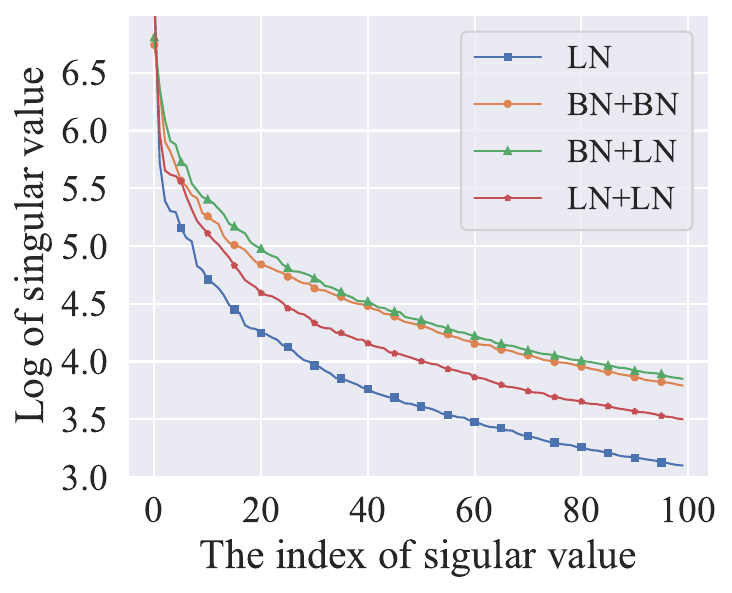} & \includegraphics[width=0.3\textwidth]{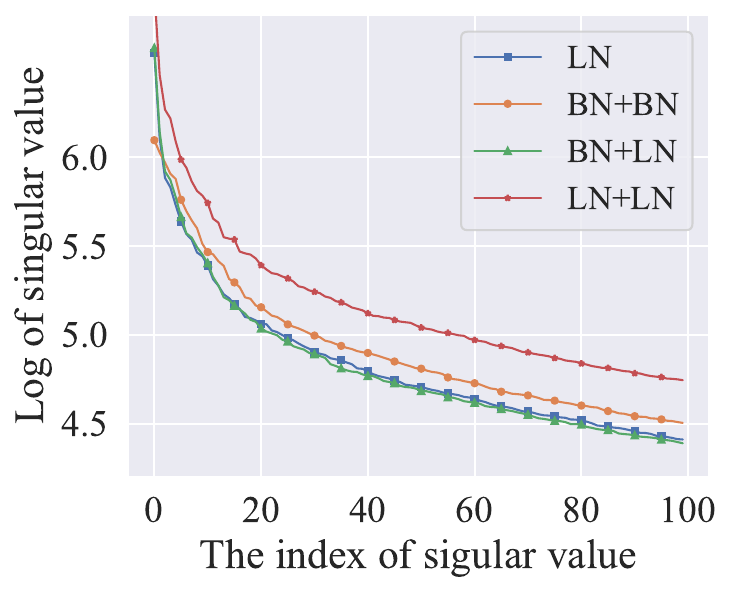}
    \\
    \quad~~(a) Pretraining loss &  \quad~~(b) \CLS embeddings &  \quad~~(c) normal token embeddings
    \end{tabular}
    \caption{
    \textbf{(a)} Reconstruction loss of the MAE pertaining -- MAE with SepNorm achieves lower MSE loss compared to ShareNorm, demonstrating a better ability to encode global contextual information. 
    \textbf{(b) \& (c)} Comparison of the singular values of learned (\CLS and normal token) features with ShareNorm and different configurations of SepNorm. \CLS embeddings learned from SepNorm have larger singular values, which suggests that vectors are better used to encode information.
    }
    \vspace{-1em}
    \label{fig:dimension_collapse}
\end{figure}

\paragraph{Experiment setup.} We pretrain MAE on the STL10 dataset via four different losses:
\vspace{-0.5em}
\begin{itemize}
    \item MAE loss $\calL_\mathrm{MAE}$ without any $\calL_{\mathrm{unif}}$ on \CLS and token embeddings. This setting is a study with MAE training only. 
    \item MAE loss $\calL_\mathrm{MAE}$ with $\calL_{\mathrm{unif}}$ on the \CLS embeddings. We treat all \CLS embeddings within the same batch (except itself) as negative instances.
    \item MAE loss $\calL_\mathrm{MAE}$ with $\calL_{\mathrm{unif}}$ on the token embeddings. We treat all token embeddings within the same batch or same images (except itself) as negative instances.
    \item MAE loss $\calL_\mathrm{MAE}$ with $\calL_{\mathrm{unif}}$ on both \CLS and token embeddings.
\end{itemize}
We choose $\lambda=\{0, 0.01,0.1,1\}$. Note that the second loss with $\lambda=0.1$ corresponds to the U-MAE~\citep{zhang2022mask}. We also replace the normalization layer of the ViT in MAE with one of the following: \{LN, BN, BN+LN, BN+BN\}. The combination of different losses, different $\lambda$'s, and different normalization layers yields 40 specifications of the experiments.

We first report our results with MAE training only. The uniformity of learned embeddings is first measured by singular values of the decomposition of an embedding matrix: we randomly choose 10k embeddings to form the matrix. We do this separately for \CLS embeddings and normal token embeddings. Figrue~\ref{fig:dimension_collapse} shows the results, which indicate that \CLS features learned from SepNorm exhibit better representational power and thus can better encode the global information.

\begin{figure}[t]
    \centering
    \small
    \begin{tabular}{cc}
        \begin{tabular}{@{}lccccc@{}}\toprule
        & \multicolumn{2}{c}{STL10} & \multicolumn{2}{c}{Aircraft} \\
        NormLayer & \CLS & token & \CLS & token\\\midrule
        LN & -2.5911 & -3.4998 & -0.9097 & -3.3514\\
        BN & -2.9531 & \textbf{-3.6785} & -1.5321 & -3.3659\\\midrule
        LN + LN & -1.1045 & -3.3685 & -1.6874 & -3.3633\\
        LN + BN & -1.3153 & -3.3735 & -1.9634 & -3.5056\\
        BN + LN &  -3.4426 & -3.3797 & -2.7705 & -3.3068 \\
        BN + BN & \textbf{-3.4763} &-3.4758 & \textbf{-2.9788} & \textbf{-3.7349} \\
        \bottomrule\\
    \end{tabular} & \begin{tabular}{c}
\includegraphics[width=0.4\textwidth]{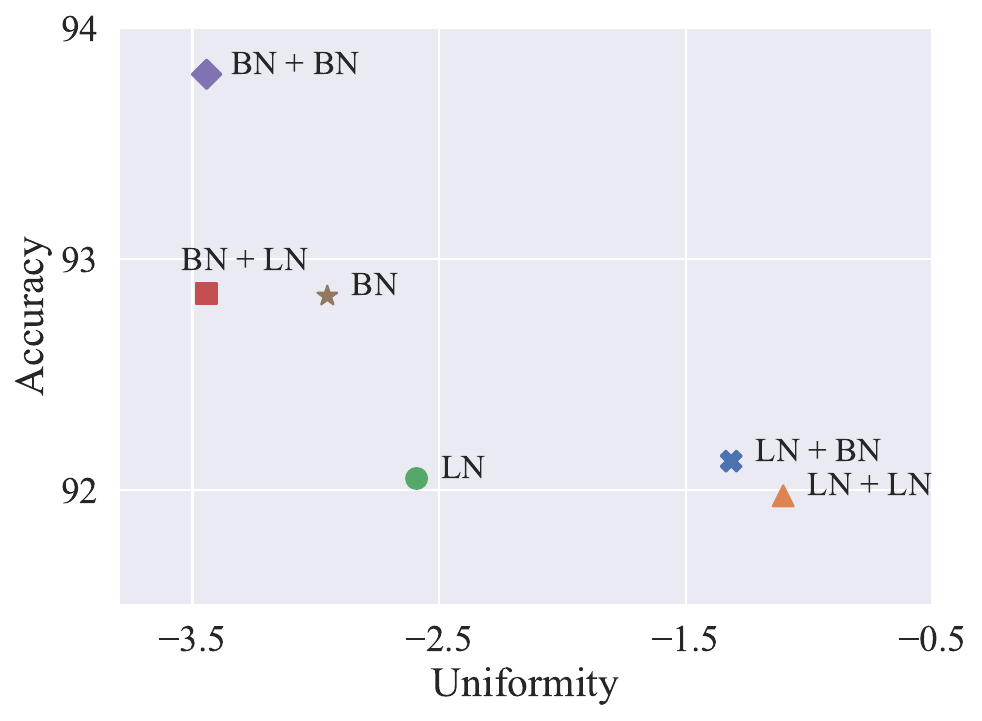}
\end{tabular}\\
(a) Uniformity of different normalization settings. & (b) Uniformity vs. accuracy
    \end{tabular}
    \caption{Uniformity Analysis. \textbf{(a)} Under SepNorm, the uniformity of the \CLS embeddings are better preserved on the STL10 and Aircraft datasets. \textbf{(b)} Uniformity is positively related to the downstream task performance -- 
    lower uniformity results in higher classification accuracy on the STL10 dataset.}
\vspace{-1em}
    \label{tab:uniformity}
\end{figure}

Then the uniformity is measured by the score in Eqn.~\ref{eq:unif}. Table~\ref{tab:uniformity}(a) shows the numerical value of the uniformity on the STL10 and Aircraft datasets~\citep{coates2011analysis, maji2013fine}. Compared to ShareNorm, \textbf{SepNorm significantly enhances the uniformity of \CLS embeddings}. Interestingly, the uniformity of normal tokens' embeddings remains comparable. We also empirically verify that better uniformity on the \CLS embeddings results in better performance on the downstream task (Figure~\ref{tab:uniformity}(b)). Another observation is that the uniformity of \CLS embeddings is clearly improved when they are normalized by BN instead of LN. Our hypothesis is that BN tries to make each feature dimension useful by controlling its variance while LN may still neglect some feature dimensions.

\begin{wrapfigure}[15]{r}{0.48\textwidth}
   \vspace{-2mm} 
    \includegraphics[width=0.46\textwidth]{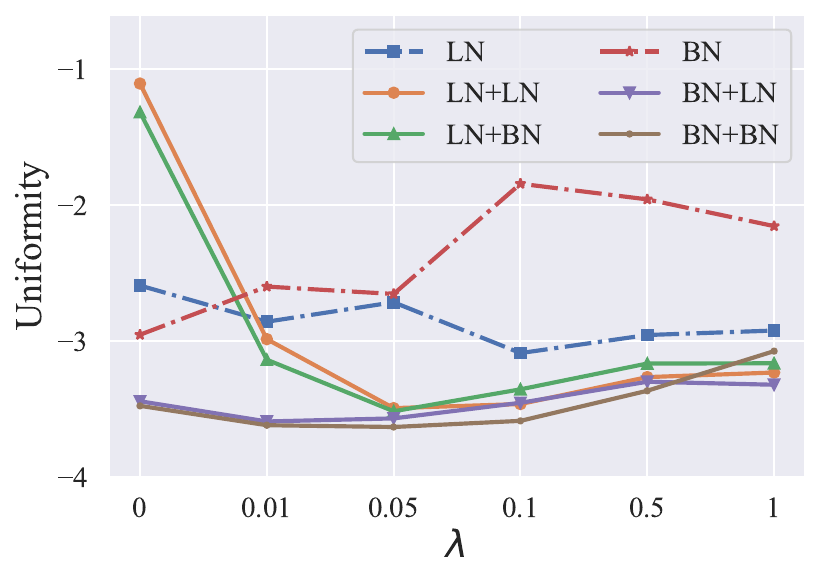}
   \vspace{-5mm} 
  \caption{$\lambda$ vs uniformity. A higher $\lambda$ emphasizes more on the uniformity of \CLS embeddings.}
  \label{fig:unif-lamb}
   \vspace{-4mm} 
\end{wrapfigure}

We then report results from studies with U-MAE training. Figure~\ref{fig:unif-lamb} shows the uniformity metrics obtained using different $\lambda$'s. When using ShareNorm, the uniformity of the \CLS embeddings is no better than -3.088, and even the explicit uniformity loss does not help much. On the contrary, embeddings learned from the proposed SepNorm can easily achieve better uniformity scores. The study with the contrastive approach further verifies the advantage of SepNorm in terms of encouraging uniformity of \CLS embeddings. 

The results provide strong evidence that \textbf{the uniformity of the \CLS embeddings is held down by ShareNorm even the minimization of an explicit contrastive loss cannot increase it}. We hypothesize that all features after LN will distribute in the same sphere, and \CLS embeddings are squeezed to a small area of the sphere surface because they need to be different from embeddings of normal tokens.

Table~\ref{tab:ablation} reports the downstream performance (accuracy) on STL10 across 40 different settings. We summarize our observations: (1) In the non-contrastive method MAE, with proper configuration, the performance of SepNorm is superior to that of  ShareNorm. (2) In contrastive methods ($\lambda \neq 0$), SepNorms' advantages are further highlighted. For example, when $\lambda=1$, the performance of SepNorm (BN+LN) is improved by 1.6\% compared to the non-contrastive method. The performance gain in SepNorm (BN+BN) is less obvious as the double BNs already impose implicit uniformity loss on both \CLS and token embeddings. 

In contrast to SepNorm, the performance of ShareNorm is less satisfactory when using contrastive methods. We believe it is very challenging to encourage the two types of embeddings to be uniformly distributed in the same sphere and keep them separable at the same time. (3) The uniformity of the token embeddings is also vital for learning a good \CLS representation, as evident by SepNorm (BN+LN) gaining accuracy with increasing $\lambda$ on the token embeddings. We hypothesize that by enforcing uniformity, the token embeddings are forced to contain less information about others, which encourages the \CLS embedding to encode as much information as possible. Our empirical study also shows that, when contrastive loss~\citep{oord2018representation} is used to encourage the uniformity of \CLS features in self-supervised transformers, the difference between BN and LN on \CLS features is not significant anymore. 

\begin{table}[t]
    \centering
    \small
    \begin{tabular}{lccccccc}
    \toprule
       {Normalization layer} & {$\lambda = 0$} 
       & {Negative pairs} 
       & {$\lambda = 0.01$} & {$\lambda = 0.1$} & {$\lambda = 1$} & Best\\
        \midrule
        \multirow{3}{*}{SepNorm (BN+BN)} & \multirow{3}{*}{93.84} & token  & 93.65 & 94.15 & 93.94 & 94.15\\
        & & \CLS &93.73 & 93.85 & 93.93 & 93.93\\ 
        && \CLS + token & 93.40 & 94.25 & 94.28 & 94.28\\\midrule
    
        \multirow{3}{*}{SepNorm (BN+LN)}  & \multirow{3}{*}{92.80}  & token & 92.98 & 93.60 & 94.30 & 94.30\\ 
        & & \CLS &92.98 & 93.48 & 93.36 & 93.48\\
        & &\CLS + token & 92.74 & 93.18 & 94.40 &\textbf{94.40}\\\midrule
        
         \multirow{3}{*}{ShareNorm (BN)}  & \multirow{3}{*}{92.84}& token& 92.48 & 93.38 & 92.78 & 93.38 \\ 
        && \CLS &93.10 & 93.33 & 92.93 &93.33\\ 
        &  &\CLS + token & 93.41 & 93.46 & 92.99 &93.46\\ \midrule
         \multirow{3}{*}{ShareNorm (LN)} & \multirow{3}{*}{92.01}&  token & 92.61 & 92.74 & 92.14 & 92.74\\
        & &  \CLS &92.28 & 92.75 & 92.36 & 92.75\\
        & &\CLS + token &  92.74 & 92.38 & 92.74 &92.74\\\bottomrule\\
    \end{tabular}
    \caption{Ablation study of the effect of \CLS and token uniformity on the downstream tasks with $\lambda$ varied. We report downstream task accuracy for the STL10 dataset.}
    \vspace{-1.5em}
    \label{tab:ablation}
\end{table}

\section{Related Works}
The training of transformer architectures with self-supervised learning has seen significant advancements in both contrastive and non-contrastive training. Among self-supervised learning methods, non-contrastive ones do not rely on negative samples for learning. They have emerged as a powerful approach for training transformer models and demonstrated remarkable successes in various tasks. BERT~\citep{devlin2018bert} and RoBERTa~\citep{liu2019roberta} were proposed in the NLP domain. Additionally, there are some works focus on the specific task, such as speech recognition~\citep{wang2020transformer}, image generation~\citep{chen2020generative}, and heterogeneous graph generation~\citep{hu2020heterogeneous}.

Contrastive methods on the contrary train networks using positive and negative samples that are constructed without manual labeling. They have also been used to train transformer-based architectures. \citet{gao2021simcse} and \citet{zhang2022mask} make significant strides in natural language processing tasks, while \citet{chen2021empirical} provide valuable insights into the pre-training of transformers. Meanwhile, the potential of contrastive methods in vision transformers has been demonstrated by~\citet{caron2021emerging} and \citet{radford2021learning}. These collective efforts underscore the versatility and efficacy of contrastive methods in self-supervised learning of transformers. 

Normalization layers, including layer normalization and batch normalization, are essential to transformer architectures because they help stabilize the training procedure and accelerate convergence. \citet{xiong2020layer} delve into the role of layer normalization in the transformer architecture and provide insights about how the layer improves the training stability and the performance of transformers. Similarly, \citet{xu2019understanding} explores the intricacies of layer normalization and offers potential enhancements to its effectiveness. To address the limitations of traditional batch normalization in a transformer architecture, \citet{shen2020powernorm} introduces a new normalization layer, Powernorm, which is a variant of batch normalization. \citet{nguyen2019transformers} focus on the normalization process in the self-attention mechanism of transformers and propose methods to optimize the normalization of self-attention. All the efforts above underscore the critical role of normalization layers in transformer models.
\vspace{-0.5em}
\section{Conclusion}
\vspace{-0.5em}
In this work, we have introduced SepNorm to separate the normalization of \CLS embeddings from that of other tokens. Across three application domains (images, text, and graphs), SepNorm shows consistent performance improvement when it is incorporated into transformer models. Our analysis shows that SepNorm promotes uniformity of \CLS embeddings and thus enhances the transformers' ability to encode information. As a valuable technique for improving the foundational transformer architecture, SepNorm has the potential to benefit a wide range of applications.
\vspace{-0.5em}
\section*{Acknowledgement}
\vspace{-0.5em}

We thank all reviewers for their constructive feedback. This research is supported by the NIGMS of the National Institutes of Health, Awards R35GM148219,  the Army Research Office, MURI program, contract \# W911NF2210239, and NSF Award 1909536. Chen and Liu are also supported by the NSF CAREER Award \# 2239869.

\bibliography{references}
\end{document}